\documentclass[times,final,authoryear]{elsarticle}
\usepackage{framed,multirow}

\usepackage{amssymb}
\usepackage{latexsym}

\usepackage{url}
\usepackage{xcolor}
\definecolor{newcolor}{rgb}{.8,.349,.1}

\journal{Journal of \LaTeX}

\newif\ifdraft
\drafttrue

\ifdraft
     \newcommand{\liu}[1]{\textcolor{blue}{{[liu: #1]}}}
\else
    \newcommand{\liu}[1]{}
\fi

\ifdraft
     \newcommand{\wm}[1]{\textcolor{magenta}{{[wwm: #1]}}}
\else
    \newcommand{\wm}[1]{}
\fi

\ifdraft
     \newcommand{\xiu}[1]{\textcolor{green}{{[xiu: #1]}}}
\else
    \newcommand{\xiu}[1]{}
\fi

\begin{document}

\ifpreprint
  \setcounter{page}{1}
\else
  \setcounter{page}{1}
\fi

\begin{frontmatter}

\title{Enhancing Local Geometry Learning for 3D Point Cloud via Decoupling Convolution}

\author[titech_es,aist_airc]{Haoyi Xiu}
\author[aist_airc]{Xin Liu \corref{cor1}}
\author[dlut]{Weimin Wang}
\author[aist_airc]{Kyoung-Sook Kim}
\author[pasco]{Takayuki Shinohara}
\author[titech_cs]{Qiong Chang}
\author[titech_es,aist_airc]{Masashi Matsuoka}

\address[titech_es]{Department of Architecture and Building Engineering, Tokyo Institute of Technology, Tokyo, Japan}
         
\address[aist_airc]{Artificial Intelligence Research Center, AIST, Tokyo, Japan}

\address[dlut]{DUT-RU International School of Information Science and Engineering, Dalian University of Technology, Dalian, China}

\address[titech_cs]{Department of Computer Science, Tokyo Institute of Technology, Tokyo, Japan}

\address[pasco]{Innovation Technology Office Research Center, PASCO Corporation, Tokyo, Japan}

\cortext[cor1]{Corresponding author: 
  e-mail: xin.liu@aist.go.jp}





\begin{abstract}
Modeling the local surface geometry is challenging in 3D point cloud understanding due to the lack of connectivity information. Most prior works model local geometry using various convolution operations. We observe that the convolution can be equivalently decomposed as a weighted combination of a local and a global component. With this observation, we explicitly decouple these two components so that the local one can be enhanced and facilitate the learning of local surface geometry. Specifically, we propose Laplacian Unit (LU), a simple yet effective architectural unit that can enhance the learning of local geometry. 
Extensive experiments demonstrate that networks equipped with LUs
achieve competitive or superior performance on typical point cloud understanding tasks. 
Moreover, through establishing connections between the mean curvature flow, a further investigation of LU based on curvatures is made to interpret the adaptive smoothing and sharpening effect of LU. The code will be available.
\end{abstract}



\end{frontmatter}



\section{Introduction}
\label{sec: introduction}
A 3D point cloud is essentially a set of points irregularly distributed on the surface of scanned objects in 3D space. The ever-growing capacity of scanning hardware enables 3D scanners to capture high-quality point clouds in a cost-effective manner. Therefore, an increasing number of point cloud datasets~\citep{geiger2012cvpr,yi2016scalable,hackel2017semantic3d,uy2019revisiting} have become available to research communities, which has triggered active research on data-driven 3D point cloud understanding for various applications such as autonomous driving~\citep{cui2021deep, qi2018frustum} and remote sensing~\citep{biasutti2019diffusion,li2022snapshotnet,zhu2017deep,shinohara2020fwnet}. 

Recently, research communities have achieved an advanced understanding of deep learning--based point cloud analysis. Compared with the representational power of conventional machine learning techniques, deep neural networks (DNNs) can learn more discriminative descriptions of data and perform exceedingly well in various research fields~\citep{lecun2015deep}. In particular, convolutional neural networks (CNNs) have shown great success in 2D image understanding~\citep{krizhevsky2012imagenet, he2016deep}, which has motivated researchers to apply CNNs to 3D point clouds.
In contrast to the regularly structured data, the unstructured nature (irregular spacing, arbitrary order, etc.) of point clouds makes the direct application of CNN challenging. Therefore, early research attempts to project point clouds onto 2D~\citep{kanezaki2018rotationnet,su2015multi} or 3D~\citep{maturana2015voxnet, zhou2018voxelnet} regular grids, thereby making the well-established convolution applicable to point clouds. However, such approaches are considered suboptimal, as they tend to lose fine geometrical details due to projections. 
To overcome this issue, PointNet~\citep{qi2017pointnet} applies pointwise multi-layer perceptrons (MLPs) and symmetric functions to the raw point clouds, successfully treating points in a lossless manner while being invariant to the point order. 

Further advancement in the DNN-based point cloud understanding is achieved by extending MLP-based methods to various local operations. PointNet++~\citep{qi2017pointnet++} applies MLPs locally to update point features using their neighbors. Subsequently, overcoming the difficulty in constructing convolution filters for unstructured points, various point convolutions are realized. Some works~\citep{hua2018pointwise,atzmon2018point,thomas2019kpconv,mao2019interpolated} explicitly introduce regular convolutional kernels to which the points are projected while the others dynamically predict the convolution filter using various features~\citep{wang2018deep,liu2019relation,li2018pointcnn,wu2019pointconv, wang2019dynamic,simonovsky2017dynamic,li2019deepgcns,liu2020closer,xu2021paconv,xiang2021walk}. Recently, inspired by the success of self-attention~\citep{vaswani2017attention}, a line of research incorporates the attention mechanism into networks~\citep{wang2019graph,zhao2019pointweb,zhao2021point,xiu2022enhancing}.

\begin{figure}[t]
    \centering
    \includegraphics[width=\linewidth]{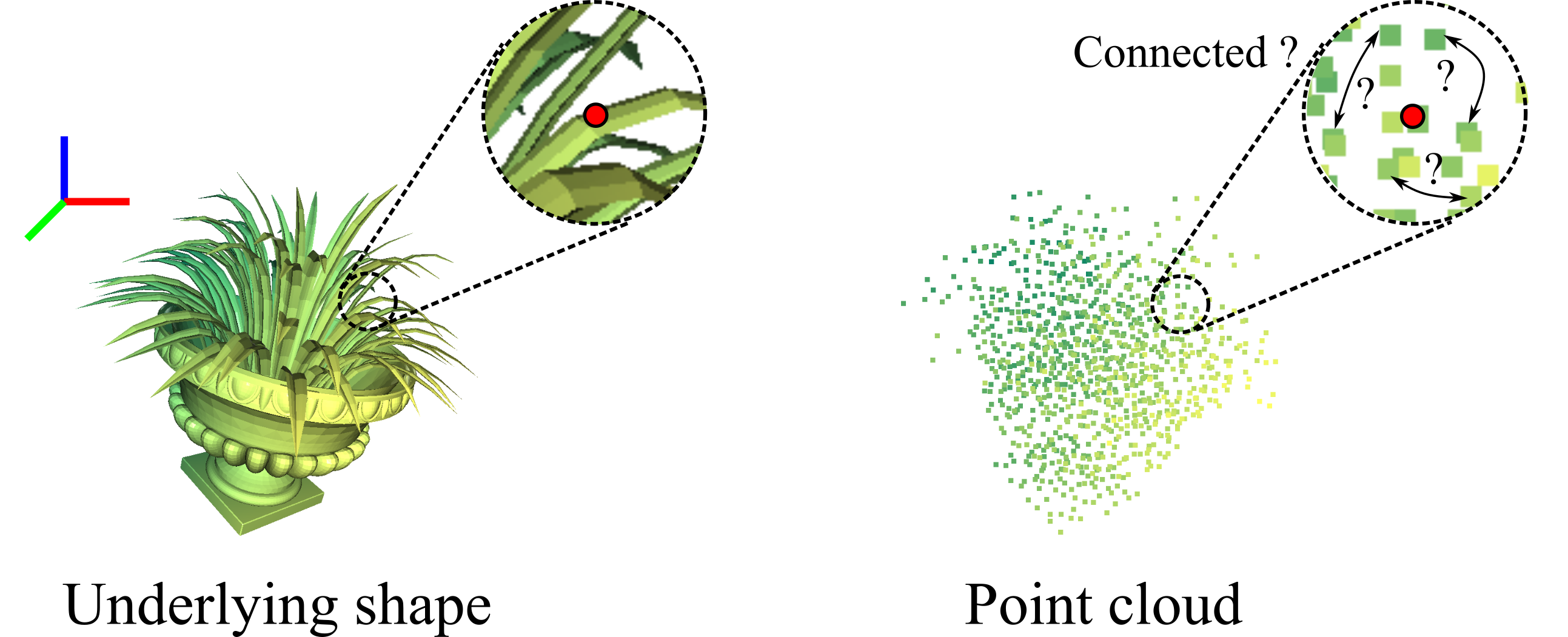}
    \caption{A point cloud lacks connectivity information, making it challenging to analyze the underlying surface geometry.}
    \label{fig: connectivity}
\end{figure}
Point clouds are irregularly distributed samples taken from object surfaces. As shown in Fig.~\ref{fig: connectivity}, unfortunately, the connectivity information is often not available; therefore, it is of great importance to model the local surface geometry of point clouds so that the obtained point representations faithfully capture the underlying surfaces.
Although Convolutional Neural Networks (CNNs) have achieved remarkable performance in various point cloud processing tasks, a careful investigation into the convolution equation reveals that: (1) the convolution consists of the modeling of local and global components, (2) the same transformations, or set of weights, are applied to both components (Sec.~\ref{sec: conv_analysis}). In other words, local and global information is always treated equally. 
This is not desirable because the two components are not of equal importance in general.
In particular, we believe that greater importance should be given to the local component due to the lack of natural connectivity between unstructured points.  

\begin{figure*}[tb]
    \centering
    \includegraphics[width=0.8\linewidth]{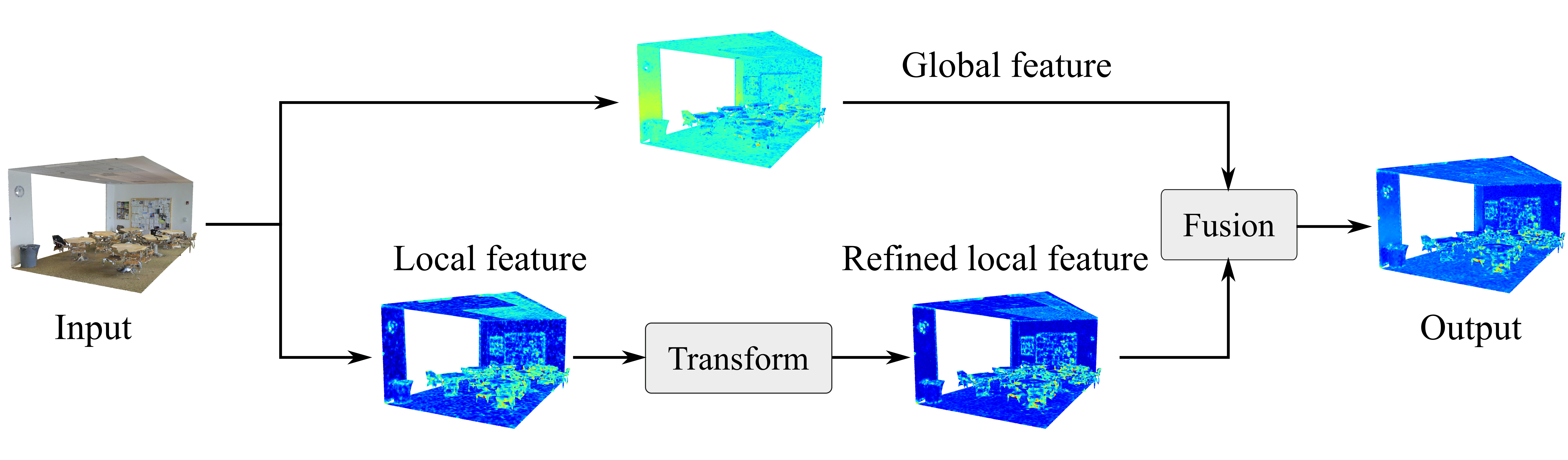}
    \caption{Overview of the LU. LU decouples the input into the global and local components where the local component is adaptively transformed, thereby facilitating the learning of local geometry. The input is shown in RGBD for visualization.}
    \label{fig: overview_LU}
\end{figure*}

In this study, to enhance the learning of local surface geometry, we propose Laplacian Unit (LU) that facilitates the modeling of local geometry by adaptively smoothing/sharpening local features. The overview of LU is shown in Fig.~\ref{fig: overview_LU}.
In particular, LU decouples the learning of local and global components that appear in the convolution equation by applying independent transformations to the local one, 
thereby providing networks the maximal flexibility to model local geometry. 

Additionally, the use of the above decoupling strategy enables us to establish a straightforward connection between LU and mean curvature flow~\citep{desbrun1999implicit}, an algorithm that smooths the surface using the curvature information. Through this connection, the behavior of LU can be intuitively understood by examining the change of curvatures. 

We further design Laplacian Unit enhanced Convolutional Neural Networks (LU-CNNs) to tackle point cloud classification, part segmentation, and scene segmentation. Through extensive experiments on challenging benchmarks, the practical effectiveness and general applicability of LU are verified by competing with recent strong networks. 

The main contributions of this work are summarized as follows:
\begin{itemize}
    \item We propose LU, a simple yet effective architectural unit dedicated to enhancing the learning of local surface geometry for point clouds.  
    \item We demonstrate the practical effectiveness and general applicability of LU through extensive experiments on several challenging benchmarks and ablation studies. In particular, the network equipped with LUs achieves state-of-the-art performance in point cloud classification while demonstrating competitive results in part and scene segmentation.   
    \item We establish a connection between LU and mean curvature flow and intuitively interpret how LU enhances local geometry learning through curvature analysis. 
\end{itemize}
\section{Related works}
\label{sec: related_works}
\subsection{Projection-based methods} 
Early attempts to apply deep learning on 3D point clouds project raw point clouds onto regular 2D (view) or 3D (voxel) grids to enable grid-based convolution operations. View-based methods~\citep{kanezaki2018rotationnet,su2015multi,feng2018gvcnn} project point clouds onto several 2D planes from different viewpoints. Generated multi-view images are subsequently processed using 2D CNNs. In contrast, voxel-based methods~\citep{maturana2015voxnet, zhou2018voxelnet,graham20183d, choy20194d} 
project point clouds onto 3D regular voxel grids and apply 3D convolutions. The performance of view-based methods relies heavily on the choice of projection planes, whereas voxel-based methods suffer from substantial memory consumption. Moreover, fine-grained geometrical details are lost due to projections.

\subsection{Point-based methods} 
Point-based methods, in contrast to projection-based methods, operate directly on raw unstructured point clouds. In particular, point clouds are naturally unordered and distributed irregularly in the 3D space; hence, such methods must be insensitive to point orders while can handle adaptively the irregular distribution. The ground-breaking work of such methods is PointNet~\citep{qi2017pointnet}, which applies shared-MLPs for embedding point features and aggregating the features by symmetric functions (e.g., max-pooling). The operations are applied to points independently and permutation-invariant, and thus the above issues are well-resolved. Following works~\citep{qi2017pointnet++,liu2020closer,zhang2019shellnet,lan2019modeling} improve the PointNet by applying PointNet-like subnetworks to local subsets of points. The overall procedure is similar to the one of the convolution layer in image processing, however, MLPs are still responsible for feature learning. 

In order to transfer the success of CNNs to point cloud processing, much effort has been invested in realizing convolution-like operations on point clouds. PointCNN~\citep{li2018pointcnn} adaptively permutes the points into the canonical order so that the standard convolution can be applied. Some methods dynamically generate filters using various features such as the relative position~\citep{wang2018deep, wu2019pointconv}, edges~\citep{simonovsky2017dynamic, wang2019dynamic,li2019deepgcns}, and combinations of features~\citep{liu2019relation, liu2020closer, xu2021paconv,xiang2021walk}. On the other hand, other approaches project point features onto the artificial kernel points. Since the kernel points have fixed order and positions, the standard convolution can be easily applied after projection. Projections are performed using methods such as trilinear interpolation~\citep{mao2019interpolated}, the Gaussian kernel~\citep{atzmon2018point, shen2018mining}, or the linear correlation~\citep{thomas2019kpconv}. On the other hand, motivated by the success of self-attention~\citep{vaswani2017attention}, attention mechanisms~\citep{vaswani2017attention} are widely adopted to dynamically compute connectivity using the point feature similarity. For instance, some works adopt the dot product for similarity measure~\citep{yang2019modeling,yan2020pointasnl} whereas edges are used in other works~\citep{wang2019graph,zhao2021point,xiu2022enhancing}. 

\section{Method}
\label{sec: laplacian_unit}
In this section, we first perform an in-depth analysis of the convolution operation from the viewpoint of local and global components. Based on the analysis, we formulate LU and subsequently provide the rationales behind its designs. Meanwhile, we further investigate LU and show the connection between LU and mean curvature flow, which enables us to interpret its behavior by examining curvatures.
Lastly, we build a family of LU-based networks, LU-CNN, for tackling various point cloud understanding tasks.     

\subsection{Analysis of the convolution operation}
\label{sec: conv_analysis}
Let $X=\{x_i\}_{i=1}^n\in\mathbb{R}^{n\times d}$ denote the feature vectors of a point cloud, where $n$ is the total number of input points, $d$ is the feature dimension, and $i$ indexes the points. 
In its simplest form, a convolution layer can be expressed as: 
\begin{equation}
\label{eq: simplified_convolution}
    x_i' = \sum_{j\in\mathcal{N}(x_i)}w_{ij} x_j,
\end{equation}
where $x_i'$ is the updated feature, $\mathcal{N}(i)$ denotes the 3D neighborhood of point $i$, and point $j\in\mathcal{N}(i)$ represents a neighbor of the point $i$ in 3D space. $w_{ij}$ is the weight associated with a neighbor $j$, which may represent a certain relationship (e.g., Euclidean distance) with the point $i$. 
Depending on the setting of $w_{ij}$, for instance, both the simple average filter or the filter as sophisticated as the bilateral filter~\citep{tomasi1998bilateral} can be expressed in the above form. 
Although Eq.~(\ref{eq: simplified_convolution}) is generally considered as the local operation, we notice that it takes as input the features that reside in the global coordinate system. In order to make this observation more explicit, Eq.~(\ref{eq: simplified_convolution}) can be alternatively rewritten as  
\begin{equation}
\label{eq: decomposed_convolution}
    x_i' = \sum_{j\in\mathcal{N}(i)}w_{ij} x_i + \sum_{j\in\mathcal{N}(i)}w_{ij} (x_j-x_i).
\end{equation}
The above decomposition shows that the convolution can be regarded as a weighted combination of a pure global feature ($\sum_{j\in\mathcal{N}(i)}w_{ij}x_i$) and a local feature ($\sum_{j\in\mathcal{N}(i)}w_{ij}(x_j - x_i)$). The global feature represents the feature of the point $i$ in the global coordinate system, while the local one describes the local surface geometry in the coordinate system centered by $i$. 

The local term of Eq.~(\ref{eq: decomposed_convolution}) is reminiscent of the discrete Laplace operator (the Laplacian)~\citep{sorkine2006differential, taubin1995signal}, which is defined as
\begin{equation}
\label{eq: diff_coords}
    \delta x_i = x_i -  \frac{1}{|\mathcal{N}(x_i)|}\sum_{j\in\mathcal{N}(x_i)}x_j,
\end{equation}
where $|\mathcal{N}(x_i)|$ denotes the number of points included in the spatial neighborhood of point $i$. 
To see it clearly, we rewrite Eq.~(\ref{eq: diff_coords}) as
\begin{equation}
\label{eq: umbrella}
    - \delta x_i = \frac{1}{|\mathcal{N}(x_i)|}\sum_{j\in\mathcal{N}(x_i)}(x_j - x_i).
\end{equation}
As a result, when $w_{ij}=\frac{1}{|\mathcal{N}(x_i)|}$ the local term of Eq.~(\ref{eq: decomposed_convolution}) is exactly the same as the Laplacian with a negative sign.

In essence, the Laplacian encodes how strong the center point deviates from the neighborhood. In other words, the operator quantifies how the surface bends around the center point. 
Such information on the local behavior is useful for characterizing the structure of an object or detecting the boundary between objects. 
Therefore, the convolution naturally takes into consideration the modeling of local surface geometry.

However, notice that in Eq.~(\ref{eq: decomposed_convolution}) the same transformation ($w_{ij}$) is applied to both local and global terms. In other words, the optimization of the local component is always coupled with the global one, forcing the two components to be treated equally.  
We believe that this is not desirable because the two components are not of equal importance in general. Consider the case of edge detection in which object boundaries are needed to be detected. The detection of boundaries is more relevant to the local characteristics than to global information like orientations. In particular for point
clouds, we consider that careful optimization of the local surface geometry is rather vital due to the lack of natural connectivity. Therefore, we propose to decouple the local component and the global component to facilitate the modeling of local surface geometry, as will be introduced in the next section.

\subsection{Laplacian Unit}
\subsubsection{Formulation}
Motivated by the above analysis, we propose LU that facilitates the modeling of local geometry by decoupling the optimization of the local and global components in the convolution.    
Specifically, we introduce transformation $\mathcal{M}\colon\mathbb{R}^{d_{in}}\to\mathbb{R}^{d_{out}}$, which is applied to the individual pairs of $x_j - x_i$:
\begin{equation}
    \label{eq: M}
    x_i' = \sum_{j\in\mathcal{N}(i)}w_{ij}x_i + \frac{1}{|\mathcal{N}(i)|}\sum_{j\in\mathcal{N}(i)}\mathcal{M}(x_j-x_i).
\end{equation}
By comparing with Eq.~(\ref{eq: decomposed_convolution}), we can consequently find that mapping $\mathcal{M}$ makes the local component no longer coupled with the global one, thereby facilitating its independent optimization. $\mathcal{M}$ filters individual channels of $x_j - x_i$ so that the useful features are enhanced while the less useful ones are suppressed. Moreover, we expect $\mathcal{M}$ in conjunction with $\frac{1}{|\mathcal{N}(i)|}$ to tackle varying density and measurement noise. In practice, we implement $\mathcal{M}$ using a single linear transformation for efficiency. Note that usually we set $d_{in}=d_{out}$ to match the input and output dimensions. Further, the fact that $\mathcal{M}$ is applied pairwise ensures that the operation is permutation-invariant; hence, it is well-suited for point cloud processing. 
Inspired by the recent practice in building DNNs, we additionally introduce a nonlinear transformation $\mathcal{T}=\texttt{ReLU}\circ\texttt{BatchNorm}$ and apply it to the local component: 
\begin{equation}
    \label{eq: T}
    x_i' = \sum_{j\in\mathcal{N}(i)}w_{ij}x_i + \mathcal{T}\left(
        \frac{1}{|\mathcal{N}(i)|}\sum_{j\in\mathcal{N}(i)}\mathcal{M}(x_j-x_i)
    \right). 
\end{equation}
$\mathcal{T}$ is introduced mainly to facilitate the training process where Batch Normalization~\citep{ioffe2015batch} is used to regularize the output while ReLU is used to encourage the sparsity~\citep{glorot2011deep}. Besides, we believe that an additional nonlinear transformation is beneficial for the further decoupling of local and global features.  

Although the learning of local and global components can be successfully decoupled with Eq.~(\ref{eq: T}), it still involves separate optimization of both components which brings new challenges for the optimal learning of the local component. In order to fully concentrate on the learning of local components, we enforce a convex combination on $w$, i.e., $\sum_{j\in\mathcal{N}(i)}w_{ij}=1$, and the term $\sum_{j\in\mathcal{N}(i)}w_{ij}x_i$ in Eq.~(\ref{eq: T}) becomes $x_i$. Consequently, Eq.~(\ref{eq: T}) can be simplified to
\begin{equation}
    \label{eq: laplacian_unit}
    x_i' = x_i + \Delta x_i,
\end{equation}
where $\Delta x_i = \mathcal{T}\left(
        \frac{1}{|\mathcal{N}(i)|}\sum_{j\in\mathcal{N}(i)}\mathcal{M}(x_j-x_i)\right)$. 
Making the learnable part only consists of the local term, Eq.~(\ref{eq: laplacian_unit}) effectively facilitates the optimization of local geometry.
Meanwhile, the above simplification transforms Eq.~(\ref{eq: laplacian_unit}) into a form of the renowned residual block~\citep{he2016deep}. Instead of learning the $x_i'$ directly, Eq.~(\ref{eq: laplacian_unit}) encourages to fit the residual mapping, i.e., $x_i' - x_i$, which greatly eases the optimization~\citep{he2016deep}. 
We name the form of Eq.~(\ref{eq: laplacian_unit}) as Laplacian Unit (LU) throughout this study. 

\subsubsection{Interpretation}
\label{sec: interpretation}
Apart from the aforementioned advantages, the form of LU also offers a convenient way to interpret its behavior. Such interpretability is valuable because it enables us to investigate how LU behaves and benefits the learning process, a trait that is often beneficial for the model design and analysis in deep learning research. 

The discrete Laplacian (Eq.~(\ref{eq: umbrella})) may be considered as an approximation of mean curvature normal~\citep{spivak1975comprehensive}. Since $\Delta x_i$ in Eq.~(\ref{eq: laplacian_unit}) can be considered as the adaptively learned discrete Laplacian, it may be approximately expressed as:
\begin{equation}
    \Delta x_i \approx H_i v_i, 
\end{equation}
where $H_i$ and $v_i$ denote the mean curvature and the unit normal vector at position $i$, respectively. In other words, $\Delta x_i$ can be expressed as a unit normal vector scaled by the mean curvature. 
Furthermore, assuming that the following relationship holds
\begin{equation}
    \partial_t x_i \approx x_i' - x_i, 
    \label{eq: forward_euler}
\end{equation}
where $t\ge0$ denotes time. Eq.~(\ref{eq: forward_euler}) is often used as the discretization scheme of the differential equations~\citep{chang2017multi}. LU then may be expressed as 
\begin{equation}
    \label{eq: mean_curvature_flow}
    \partial_t x_i = H_i v_i, 
\end{equation}
which is identical to the definition of the mean curvature flow~\citep{desbrun1999implicit}. Mean curvature flow smooths the surface by deforming the surface along the direction of the normal vector with a speed proportional to the mean curvature. The surface evolves under the flow continues to become smoother as the time (or the number of iterations in the discrete sense) goes on. Therefore, it is reasonable to assume that LU behaves like mean curvature flow and smooths out small variations in practice. 

Notice that, however, $\Delta x_i$ is adaptively learned; therefore, LU 
is able to perform smoothing as well as sharpening (i.e., the inverse of smoothing) by changing the direction of the vector. Since the over-smoothing problem is likely to occur in CNNs~\citep{li2018deeper}, such an adaptivity is useful for preventing it. 

To shed light on the underlying mechanism of LU, we can evaluate the change of the mean curvature before and after applying LU. Specifically, the increased curvature implies that the local surface undergoes a sharpening while the curvature becomes small when it is smoothed. 
Details of the analysis are presented in Sec.~\ref{sec: visual_interpretation}.  
\begin{figure*}[tb]
    \centering
    \includegraphics[width=0.9\linewidth]{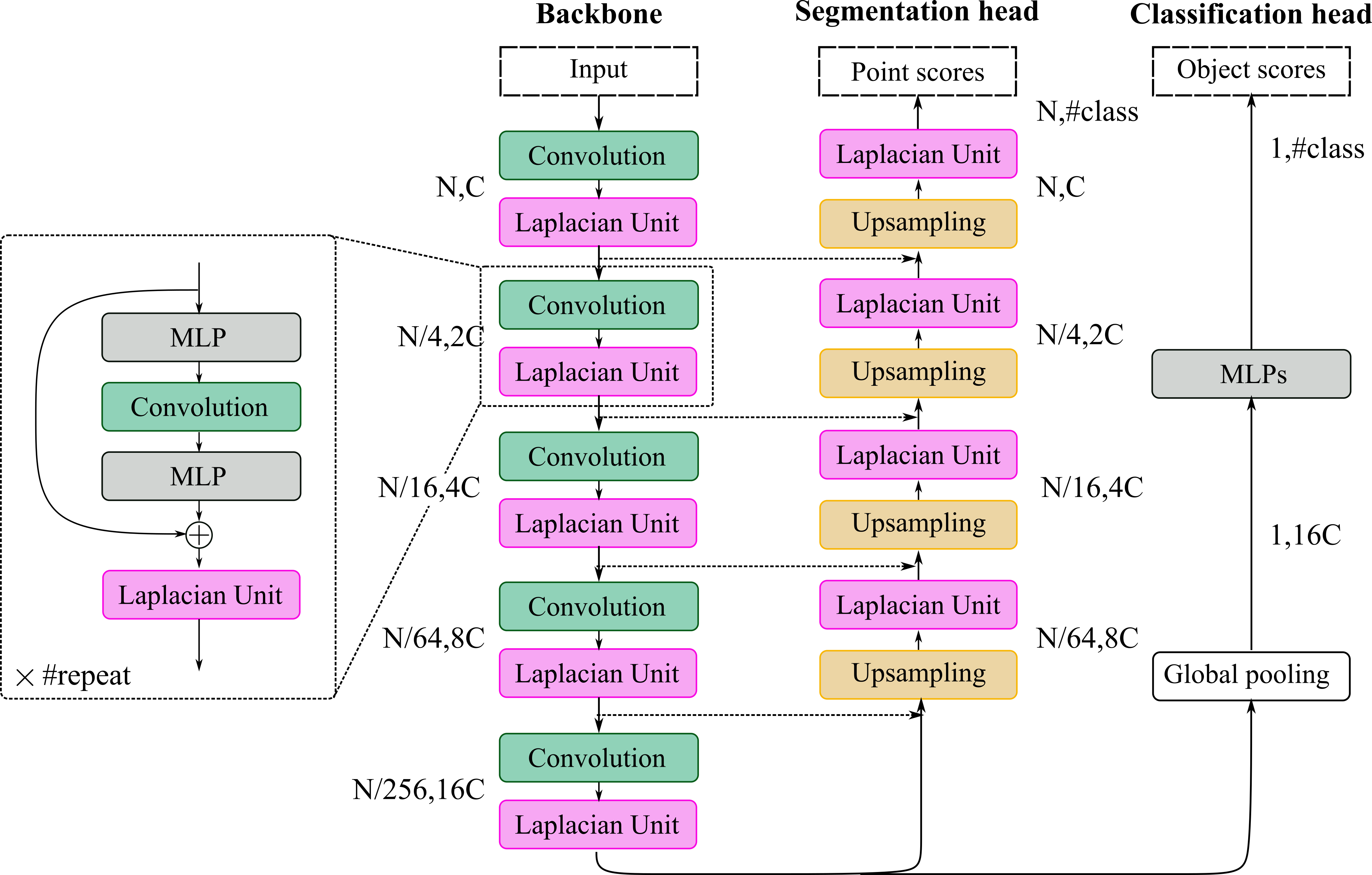}
    \caption{Architectures of LU-CNN for various point cloud understanding tasks. In general, any point cloud convolution method can be placed into the convolution block. In this work, we develop KPConv-DS, an efficient variant of KPConv with the depthwise separable convolution strategy, as the convolution method. The dashed box illustrates the memory-efficient bottleneck design~\citep{he2016deep}.}
    \label{fig: architecture}
\end{figure*}
\subsection{LU-CNN}
\label{sec: LU-CNN}

In this section, we construct LU-CNN, a powerful family of models for tackling various point cloud understanding tasks. The design of LU-CNN is determined by three major elements: LU, the convolution block, and the network architecture. 
First, the convolution method used to build the convolution block is described.  
Then, we elaborate on how the LUs and convolutions are arranged to form network architectures that tackle point cloud classification and segmentation. The overview of the architecture is presented in Fig.~\ref{fig: architecture}.


\paragraph{Convolution method}
CNN has been the most effective network architecture for the point cloud recognition tasks. The core of CNN is the convolution operation that performs the major part of feature learning. Among various point convolutions, we choose KPConv~\citep{thomas2019kpconv} as our basic convolution method for its outstanding performance and general applicability. However, like the standard convolution, KPConv causes much memory consumption, limiting the construction of deep/wide models; hence, we design an efficient variant of KPConv that enjoys lower memory consumption without compromising performance. 
Concretely, we follow the well-known depthwise separable (DS) framework~\citep{sifre2014rigid,howard2017mobilenets,chollet2017xception,sandler2018mobilenetv2} to simplify the standard convolution operation into depthwise one, drastically reducing the memory consumption. To maintain the expressiveness, we augment the input features by additionally concatenating relative positions and 3D Euclidean distance to the input feature following~\citep{liu2019relation, xu2021paconv}. The augmented features are transformed by MLPs and subsequently aggregated by the depthwise convolution. 
We denote the resulting efficient convolution method as KPConv-DS throughout this study. We adopt KPConv-DS as the basic convolution method for LU-CNN.   
 
\paragraph{Network architectures}
Two network architectures are developed for classification and segmentation. Both networks have a similar five-stage encoder where each stage corresponds to a resolution. In each stage, point features are transformed by consecutive applications of convolution blocks and LUs. In particular, we adopt the bottleneck design~\citep{he2016deep} (illustrated in the dashed box in Fig.~\ref{fig: architecture}) in which input feature dimensions are reduced before and restored after the convolution. The bottleneck design enables LU-CNN to reduce memory consumption without compromising performance. The dimension reduction and restoration are implemented using an MLP. 

LUs are applied to each stage to exploit the resolution-dependent local characteristics of objects. Note that each object is expected to have its own ``favorite'' resolution; thus, it is challenging to select a specific resolution that may bring the maximal performance gain without trial and error. Owing to its lightweight nature, LU can be easily applied to each stage without exceeding computational overhead and optimization difficulty. Points are downsampled when transitioning to the next stage so that hierarchical representations can be efficiently encoded. Among many downsampling algorithms, we adopt the furthest point sampling~\citep{qi2017pointnet++} to ensure that points are sampled uniformly. 

For the classification task, the encoder is followed by the classification head that aggregates features into a global representation by the global average pooling. The global feature vector is then transformed by a series of MLPs to produce class scores for the input object. The design of the classification head is illustrated in Fig.~\ref{fig: architecture}.

For the segmentation task, the output of the encoder is fed into the segmentation head. The segmentation head gradually upsamples the points using trilinear interpolation until it recovers the full resolution. During each upsampling, the U-Net~\citep{ronneberger2015u} style skip connections are used to assist the feature reconstruction. Similar to the encoder, LUs are applied after each upsampling layer so that the features are appropriately smoothed/sharpened after interpolation. Subsequently, the features are transformed by MLPs to produce point-wise scores. The overview of the segmentation head is illustrated in Fig.~\ref{fig: architecture}.

\section{Experiments}
In this section, we report the result of experiments performed on several challenging benchmarks. 
First, we report the performance of LU-CNN on point cloud classification, part segmentation, and scene segmentation. 
Next, the result of ablation studies is reported and analyzed. 
Then, we analyze the additional computational cost caused by LUs. Lastly, we visually demonstrate the result of curvature analysis by which we intuitively investigate the behavior of LU.

All experiments are performed using PyTorch deep learning framework on a server with four NVIDIA V100 GPUs. The classification and part segmentation models are trained using a GPU whereas the scene segmentation model is trained using four GPUs. The kNN algorithm is used for the neighborhood construction in the experiments.

\subsection{Classification}

\paragraph{Dataset} 
We use the ScanObjectNN dataset to evaluate the performance of LU-CNN on point cloud classification. ScanObjectNN consists of 15k common objects (e.g., chairs and desks) which are collected from real-world 3D scans. There are 15 classes in total and each object is categorized into one of the 15 classes. A single point cloud of an object contains 2,048 points. As they are real-world scans, each point cloud includes measurement errors, certain occlusions, varying densities, and background points. We use the hardest train-test set~\citep{uy2019revisiting}, where objects are randomly perturbed, translated, and rotated, and adopt the official train-test split, where 80\% of the data are used for training and the remaining 20\% for the test. 

\paragraph{Setting} 
We use the SGD optimizer and trained the model for 150 epochs. The initial learning rate is set to 0.1 and decayed by a factor of 10 when the number of epochs reaches 90 and 120 epochs. 
Like previous works, we use 1,024 points as input.
Each input is normalized such that the maximum spatial distance from the origin to a point is 1. We apply random rotation, random translation, and random anisotropic scaling for data augmentation. The batch size is set to 24. Following prior works, overall accuracy (OA) is used to measure performance.

\paragraph{Result}
The quantitative result is shown in Table~\ref{tab: result_cls}. LU-CNN successfully achieves the best performance among recent powerful networks. 
Notice that our plain network without LUs performs on par with the recently proposed PointMLP~\citep{ma2022rethinking}, which uses a similar residual architecture as our network, verifying the effectiveness of our backbone network. 
With the help of LUs, the performance of our backbone further improves and successfully achieves the state of the art. We believe that LU manages to smooth out local small variations while salient edges can be reliably detected, thus leading to improved performance. One might observe that the effect of LU is not as significant as in more challenging part segmentation (Sec.~\ref{sec: partseg}) and scene segmentation tasks (Sec.~\ref{sec: scene_seg}). We conjecture that the enhancement of local information is less crucial in classification than in segmentation because the final scores are produced by a globally averaged feature vector.  

\begin{table}[t]
    \small
    \caption{Result of object classification. The bold number shows the best performance.}
    \begin{center}
        \begin{tabular}{l|c}
        \hline
        Method
        & OA
        \\
        \hline
        PointNet~\citep{qi2017pointnet}
        & 68.2
        \\
        PointNet++~\citep{qi2017pointnet++}
        & 77.9
        \\
        PointCNN~\citep{li2018pointcnn}
        & 78.5
        \\
        DGCNN~\citep{wang2019dynamic}
        & 78.1
        \\
        BGA-PN++~\citep{uy2019revisiting}
        & 80.2
        \\
        BGA-DGCNN~\citep{uy2019revisiting}
        & 79.7
        \\
        SimpleView~\citep{goyal2021revisiting}
        & 80.5
        
        \\
        GBNet~\citep{qiu2021geometric}
        & 80.5
        \\
        DynamicScale~\citep{sheshappanavar2021dynamic}
        & 82.0
        \\
        MVTN~\citep{hamdi2021mvtn}
        & 82.8
        \\
        PointMLP~\citep{ma2022rethinking}
        & 86.1
        \\
        \hline
        Ours (w/o LU)
        & 86.1 
        \\
        Ours 
        & \textbf{86.2}
        \\
        \hline
        \end{tabular}
    \end{center}

\label{tab: result_cls}
\end{table}

\subsection{Part Segmentation}
\label{sec: partseg}

\begin{table}[t]
    \small
    \caption{Result of object part segmentation. The bold numbers indicate the best performance.}
    \begin{center}
        \begin{tabular}{l|cc}
        \hline
            Method
            & ImIoU
            & CmIoU
            \\
            \hline
            PointNet++~\citep{qi2017pointnet++}
            & 85.1
            & 81.9
            \\
            PointCNN~\citep{li2018pointcnn}
            & 86.1
            & 84.6
            \\
            DGCNN~\citep{wang2019dynamic}
            & 85.2
            & 82.3
            \\
            PointConv~\citep{wu2019pointconv}
            & 85.7
            & 82.8
            \\
            
            KPConv~\citep{thomas2019kpconv}
            & 86.2
            & 85.0
            \\
            
            KPConv deform~\citep{thomas2019kpconv}
            & 86.4
            & 85.1
            \\
            
            PCT~\citep{guo2021pct}
            & 86.4
            & 83.1
            \\
            Point Transformer~\citep{zhao2021point}
            & 86.6
            & 83.7
            \\
            CurveNet~\citep{xiang2021walk}
            & 86.8
            & -
            \\
            PAConv~\citep{xu2021paconv}
            & 86.1
            & 84.9
            \\
            AGCN (No adv.)~\citep{kim2021agcn}
            & 86.3
            & 84.4
            \\
            AGCN (Full)~\citep{kim2021agcn}
            & \textbf{87.9}
            & \textbf{86.7}
            \\
            PointMLP~\citep{ma2022rethinking}
            & 86.1
            & 84.6
            \\
            \hline
            Ours (w/o LU)
            & 86.8 
            & 84.2 
            \\
            Ours
            & 87.2 
            & 84.9 
            \\
            \hline
        \end{tabular}
    \end{center}

\label{tab: result_shapenet}
\end{table}

\begin{figure}[tb]
    \centering
    \includegraphics[width=1.0\linewidth]{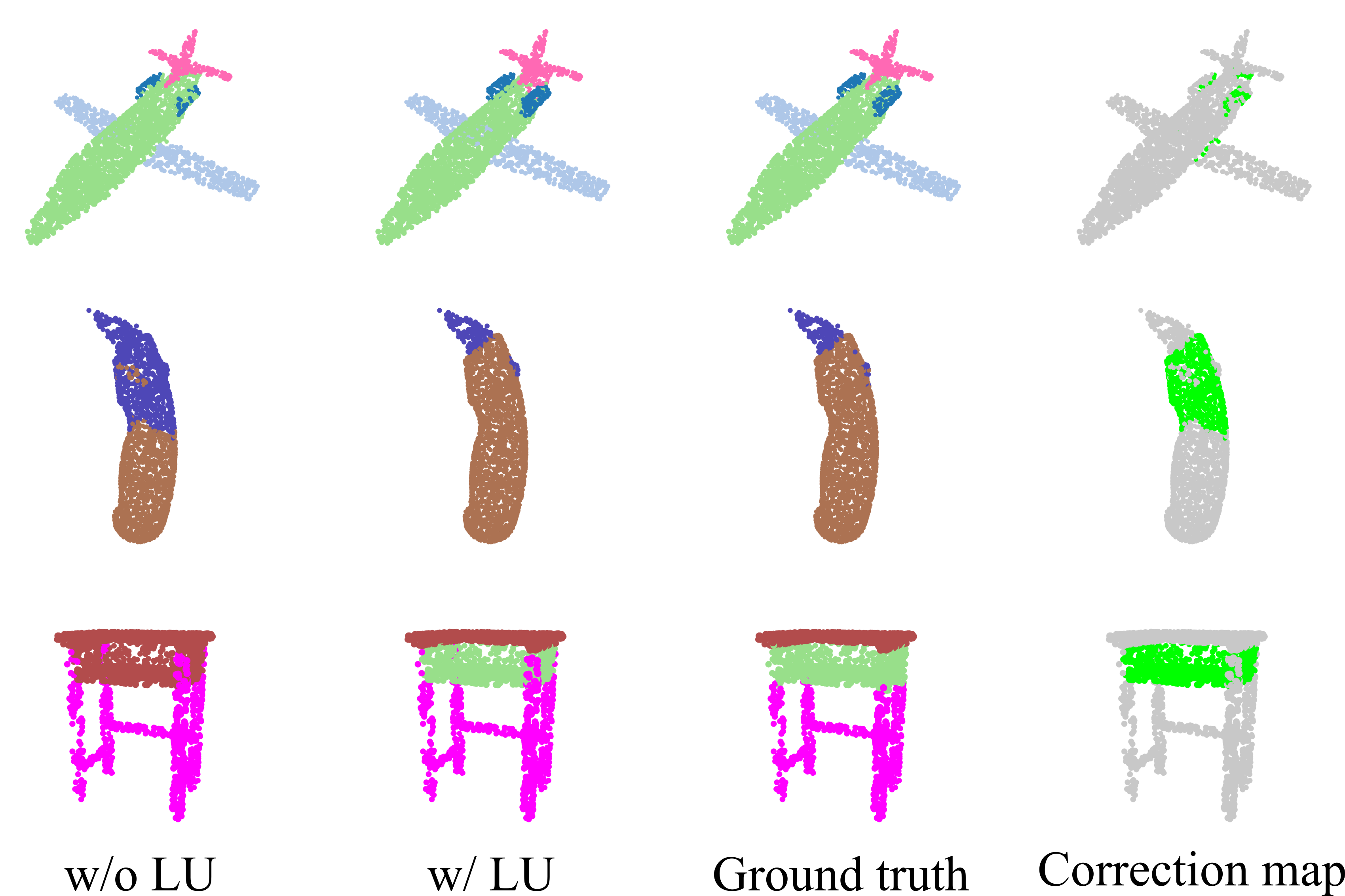}
    \caption{Qualitative result of part segmentation. The correction map indicates the predictions that are corrected by combining LUs with the plain network. \textcolor{green}{Green} points show the corrected prediction. Compared with the plain network (w/o LU), LU-CNN (w/ LU) makes the network more sensitive to part boundaries, which prevents over-smoothing and results in more perceptually sound predictions.}
    \label{fig: result_shapenet}
\end{figure}
\paragraph{Dataset}
We adopt widely used ShapeNet Part~\citep{yi2016scalable} dataset for part segmentation. This dataset contains 16,880 synthetic 3D objects. Categories included are some common objects like the hat and knife. It contains a total of 16 object categories with 50 part categories where each object is annotated into two to six parts. Each point cloud contains around 2,300 points. For benchmarking purpose, we use the data provided by \citep{qi2017pointnet++}. The standard train-test split in which 14,006 models are used for training and 2,874 models for testing is adopted.

\paragraph{Setting}
For this task, we use all available points (average 2,300 points for each point cloud) with their surface normal features as input. Like in the classification, each cloud is normalized to fill the unit ball. The input data are augmented by random anisotropic scaling and random translation. We train the models for 150 epochs using the SGD optimizer. The initial learning rate is set to 0.1 which is decayed by a factor of 10 when it reaches 90 and 120 epochs. The batch size is set to 32. Following the common procedure, we perform the voting post-processing~\citep{thomas2019kpconv, xu2021paconv} to measure the test performance. Following prior works, Instance-wise average intersection over union (ImIoU) and category-wise average IoU (CmIoU) are used for performance assessments. Both metrics are calculated according to \citep{qi2017pointnet,wang2019dynamic}.
Regarding the use of object category labels (notice that we predict part categories), we also follow the common procedure~\citep{qi2017pointnet,wang2019dynamic} by treating it as an additional one-hot feature vector.  

\paragraph{Result}
The results are reported in Table~\ref{tab: result_shapenet}. As can be seen, LU-CNN achieves competitive performance compared with recent works. 
By applying LUs to our plain network, the performance is significantly improved by 0.4 ImIoU and 0.7 CmIoU points, successfully demonstrating the effectiveness of LU. 
As shown in the first row of Fig.~\ref{fig: result_shapenet}, LU-CNN produces more precise predictions in near boundary regions compared to the plain counterpart. Therefore, LUs are especially effective in locating object boundaries. This point is also verified in the second row of Fig.~\ref{fig: result_shapenet} in which LU-CNN manages to detect very subtle changes of the surface and prevents over-smoothing. In contrast to the plain network which is confused by similar parts (e.g., tabletop and drawer, the third row of Fig.~\ref{fig: result_shapenet}), LU-CNN clearly identifies two parts as separated ones and produces more perceptually sound predictions. We conjecture that the recognition of boundaries helps the network to separate different parts.          
Although LU-CNN fails to outperform AGCN~\citep{kim2021agcn} that adopts additional adversarial training, we observe that our network beats the plain AGCN (AGCN (No adv.) in Table~\ref{tab: result_shapenet}) that trained under the similar setting as ours significantly. Therefore, we believe that LU-CNN is fairly competitive among recent strong methods under the similar training setting.

\subsection{Scene Segmentation}
\label{sec: scene_seg}

\paragraph{Dataset}
We use Stanford Large-Scale 3D Indoor Spaces (S3DIS)~\citep{armeni20163d} for scene segmentation. In total, six indoor environments (areas) containing 272 rooms are included. Each point is labeled with a class from 13 categories. The number of points contained in a room ranges from 0.2M to 4.5M. As suggested by \citep{tchapmi2017segcloud}, we use Area-5 for testing and others for training. 

\paragraph{Setting}
Since most rooms contain over 1M points which
are difficult to fit into the GPU memory, we voxelize and downsample each room with a resolution of 0.04 m. As a result, each mini-batch element consists of a point cloud that contains at most 80,000 points. During testing, we make sure that every point is evaluated.
The input feature consists of 3D coordinates and colors. We use the SGD optimizer with an initial learning rate of 0.1. The learning rate is decayed by a factor of 10 when the number of epochs reaches 60 and 80. The model is trained for 100 epochs in total. Since we can crop arbitrary numbers of training samples from rooms, we manually set the number of iterations in each epoch to 400. The batch size is set to 16.
We augment the input data using random anisotropic scaling, random color translation, color jittering, and color translation in HSV space. 
Following prior works, we assess the performance using the point average IoU (mIoU), overall accuracy (OA), and mean accuracy (MA).

\begin{table}[tb]
\small
    \caption{Result of indoor scene segmentation. The bold numbers indicate the best performance.}
    \begin{center}
        \begin{tabular}{l|cccc}
        \hline
            Method
            & mIoU
            & OA
            & MA
            \\
            \hline
            PointCNN~\citep{li2018pointcnn}
            & 57.3 
            & 85.9
            & 63.9
            \\
            KPConv~\citep{thomas2019kpconv}
            & 65.4
            & -
            & 70.9
            \\
            KPConv deform~\citep{thomas2019kpconv}
            & 67.1
            & -
            & 72.8
            \\
            PointWeb~\citep{zhao2019pointweb}
            & 60.3
            & 87.0
            & 66.6
            \\
            Minkowski~\citep{choy20194d}
            & 65.4
            & -
            & 71.7
            \\
            BAAF-Net~\citep{qiu2021semantic}
            & 65.4
            & 88.9
            & 73.1
            \\
            PAConv~\citep{xu2021paconv}
            & 66.6
            &-
            &-
            \\
            CGA-Net~\citep{lu2021cga}
            & 68.6
            & - 
            & -
            \\
            RFCR~\citep{gong2021omni}
            & 68.7
            & -
            & -
            \\
            Point Transformer~\citep{zhao2021point}
            & \textbf{70.4}
            & \textbf{90.8}
            & \textbf{76.5}
            \\
            \hline
            Ours (w/o LU)
            & 68.3 
            & 90.5 
            & 74.8 
            \\
            Ours
            & 69.6
            & 90.7 
            & 75.9 
            \\
            \hline
        \end{tabular}
    \end{center}

\label{tab: result_indoor}
\end{table}
\begin{figure}[tb]
    \centering
    \includegraphics[width=1.0\linewidth]{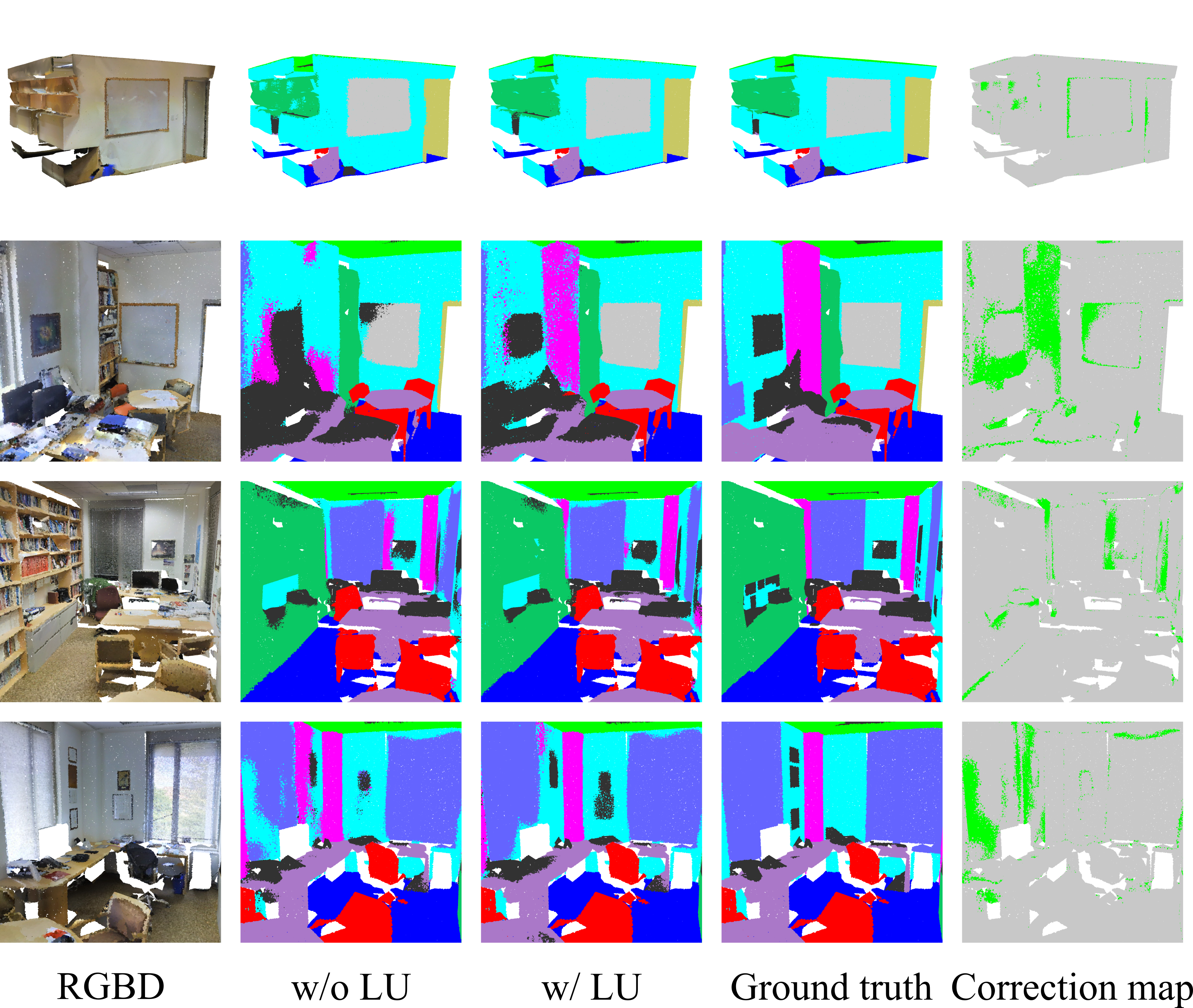}
    \caption{Qualitative result of scene segmentation. The correction map indicates the predictions that are corrected by combining LUs with the plain network. \textcolor{green}{Green} points show the corrected prediction. Owing to its ability to perform smoothing and sharpening adaptively, LU encourages the network to produce sharp boundaries while keeping within-boundary regions smooth. }
    \label{fig: result_s3dis}
\end{figure}

\paragraph{Result}
The result of scene segmentation is reported in Table~\ref{tab: result_indoor}. LU-CNN achieves the second-best performance among recent strong methods.  
Although our plain network fails to compete with the recent cutting-edge networks, LUs significantly advance its performance by 1.3 mIoU, 0.2 OA, and 1.1 MA points, respectively. 
The primary effect of LU is the accurate localization of object boundaries. As can be seen in the correction map of the first row of Fig.~\ref{fig: result_s3dis}, the improvement is shown in near-boundary areas, making predictions more faithful to the ground truth. 
Secondly, owing to its ability to perform adaptive smoothing, the network with LUs, in general, produces smoother predictions. For instance, the fourth row of Fig.~\ref{fig: result_s3dis} shows that not only the boundary of the window is recognized precisely but also the predictions of the within-boundary region present a smoother distribution compared to the one without LUs. Furthermore, with the increased sensitivity to the geometrical/semantic changes of the surface, the network with LUs detects objects that are completely ignored by the plain counterpart and considerably improves the predictions both qualitatively and quantitatively (e.g., the second row of Fig.~\ref{fig: result_s3dis}). 
We believe that giving the freedom to perform smoothing or sharpening to LU makes the network more aware of the connectivity of the underlying surface, thus making predictions smoother within and sharper near object boundaries.

\subsection{Ablation study}
\label{sec: ablation}
In this section, a wide range of experiments are conducted to investigate the design choices of LU. Specifically, we perform the component analysis to validate the influence of each component w.r.t the final performance. Next, the generalization ability of LU with regard to different local operators is assessed. All experiments are performed on the part segmentation task because we believe that the effect of adaptive local feature learning is significant in this task as analyzed in Sec.~\ref{sec: partseg}. 
\subsubsection{Component analysis}
\paragraph{Transformations $\mathcal{M}$ and $\mathcal{T}$} 
Both transformations together transform the raw discrete Laplacian into the learnable one as defined in Eq.~(\ref{eq: T}). In other words, they are responsible for adaptively smoothing or sharpening the features so as to improve the performance. As shown in Table~\ref{tab: ablation}, the networks (Model B and C) achieve degraded performance when either of them is removed. Notice that the Model B and C still able to improve over the baseline; thus, each function has a favorable effect on the overall performance. Removing both mappings (Model D) obtain significantly reduced performance whereas the full network (Model A) outperforms the Model B and C, demonstrating that both components work jointly to achieve the best performance.
\begin{table}[t]
    \centering
    \caption{Results of the ablation study on ShapeNet Part dataset. We investigate the effect of transformations $\mathcal{M}$ and $\mathcal{T}$, fusion methods of local and global features, and the number of neighbor points. Here, the performance of model A is different from the one reported in the Table~\ref{tab: result_shapenet} because we do not perform the voting post-processing for ablation experiments.}
    \begin{tabular}{c|ccccc}
        \hline
        Model
        & $\mathcal{M}$
        & $\mathcal{T}$
        & Fusion
        & $|\mathcal{N}(i)|$
        & ImIoU
        \\\hline
        Baseline
        & -
        & -
        & -
        & -
        & 86.1
        \\
        A 
        & \checkmark
        & \checkmark
        & Add.
        & 16
        & 86.7
        \\
        B
        & 
        & \checkmark
        & Add.
        & 16
        & 86.4
        \\
        C
        & \checkmark
        & 
        & Add.
        & 16
        & 86.4
        \\
        D
        & 
        & 
        & Add.
        & 16
        & 85.5
        \\
        E
        & \checkmark
        & \checkmark
        & Concat.
        & 16
        & 86.5
        \\
        F
        & \checkmark
        & \checkmark
        & Mul.
        & 16
        & 68.1
        \\
        G
        & \checkmark
        & \checkmark
        & 
        & 16
        & 85.4
        \\
        H
        & \checkmark
        & \checkmark
        & Add.
        & 8
        & 86.5
        \\
        I
        & \checkmark
        & \checkmark
        & Add.
        & 24
        & 86.5
        \\
        J
        & \checkmark
        & \checkmark
        & Add.
        & 32
        & 86.7
        \\\hline
        
    \end{tabular}
    
    \label{tab: ablation}
\end{table}

\paragraph{Local-global fusion method}
To combine the local feature and the global feature, we use the addition (Add.) fusion in LU by default because the addition naturally appears in the decomposed filtering equation (Eq.~(\ref{eq: decomposed_convolution})). Moreover, addition also makes the relationship between LU and mean curvature flow straightforward (Sec.~\ref{sec: interpretation}). However, one may conjecture that other fusion methods are also effective. Therefore, we explore two popular fusion methods: the concatenation (Concat.)~\citep{wang2019dynamic} (Model E) and multiplication (Mul.)~\citep{hu2018squeeze} (Model F). Additionally, we also present the result without any fusion, i.e., $\Delta x_i$ in Eq.~(\ref{eq: laplacian_unit}) is treated as the output of LU. 
The results are shown in Table~\ref{tab: ablation}. Obviously, the Add. is a particular solution of the Concat., and thus the Concat. should at least be not inferior to Add.. We find that, however, Concat. degrades the performance by 0.2, which contradicts our intuition. The reason for this might be due to the difficulty of joint optimization of local and global components. Moreover, we believe that it is challenging for the solver to exactly approximate the Add.. Further, the more expressive Concat. falls short of simple Add. proves that our design choice is more effective. On the other hand, Mul. degrades the performance significantly. Mul. fusion multiplies the global and local representations, which makes the backpropagated gradients for local and global features be tied together. We believe that too many interactions between two representations during optimization can increase the optimization difficulty as they represent fairly different properties of a point. 

\paragraph{Number of neighbors $|\mathcal{N}(i)|$} 
We vary the number of neighbors involved in LU to investigate its impact. $|\mathcal{N}(i)|$ is varied from 8 to 32. The results are shown in Table~\ref{tab: ablation}. The best performance are reached when $|\mathcal{N}(i)|$ is 16 (Model A) or 32 (Model J) whereas the performance drop when $|\mathcal{N}(i)|$ is set to 8 (Model H) or 24 (Model I). Therefore, we adopt 16 as the default choice.

\subsubsection{LU with various local operators}

\begin{table}[t]
    \centering
    \caption{Impact of LU on different networks. A different network is constructed by replacing the default convolution operator (KPConv-DS) in our architecture with a different local operator. The impact is measured on ShapeNet Part dataset. The performance of the KPConv-DS is different from the ones reported in Table~\ref{tab: result_shapenet} because we do not perform the voting post-processing for ablation experiments.  }
    \begin{tabular}{l|ccc}
         \hline
         Local operator
         & w/o LU
         & w/ LU
         & $\bigtriangleup$
         \\\hline
         PointNet++~\citep{qi2017pointnet++}
         & 85.1
         & 85.5
         & +0.4
         \\
         PointConv~\citep{wu2019pointconv}
         & 86.3
         & 86.7
         & +0.4
         \\
         RSCNN~\citep{liu2019relation}
         & 85.9
         & 86.2
         & +0.3
         \\
         KPConv~\citep{thomas2019kpconv}
         & 86.2
         & 86.3
         & +0.1
         \\
         \hline
         KPConv-DS (Ours)
         & 86.1
         & 86.7
         & +0.6
         \\\hline
    \end{tabular}
    
    \label{tab: conv_op}
\end{table}
We explore the impact of LU on different networks by varying the local operators. Specifically, we fix our backbone architecture and replace the KPConv-DS with other local operators to construct different networks. Four widely used local operators are chosen and evaluated. Then, we compare the performance before and after adding LUs to architectures. The results are reported in Table~\ref{tab: conv_op}. 
Results show that LU can consistently provide performance improvement for different networks, demonstrating its general applicability. Notably, KPConv-DS (Ours) achieves similar performance as KPConv~\citep{thomas2019kpconv}, showing the effectiveness of our modifications on KPConv.  

\subsection{Computational complexity}
\begin{table}[t]
    \centering
    \caption{Additional computational complexity measured on the ShapeNet Part dataset. \#param. indicates the number of parameters (M) while Speed shows the inference speed (ms) per mini-batch. }
    \begin{tabular}{l|ccccc}
    \hline
         \#LU
         & No
         & 2
         & 4
         & 6
         & Full (18)
         \\\hline
         \#param.(M)
         & 6.76
         & 6.78
         & 6.82
         & 6.95
         & 11.33
         \\
         Speed (ms)
         & 22.9
         & 23.7
         & 24.3
         & 24.8
         & 27.4
         \\
         ImIoU
         & 86.1
         & 86.3
         & 86.5
         & 86.5
         & 86.7
         \\\hline
    \end{tabular}
    
    \label{tab: complexity}
\end{table}
In this section, we analyze the additional parameters and inference time caused by combining LUs on part segmentation. In particular, we gradually increase the number of LU in the LU-CNN by adding a pair of LUs to the end of each stage. For instance, when the number of LU is set to two, a pair of LUs are added to the end of stage one in the encoder and the decoder simultaneously. The number of LU is then gradually increased from 0 to full to measure accuracy-complexity trade-offs. The results are listed in Table~\ref{tab: complexity}. As can be seen, the increase of parameters and the inference time introduced by LUs are marginal when \#LU. is set to 2--6. On the other hand, the performance approaches the one of the Full model (86.7) quickly as it achieves 86.5 when \#LU. is only 6. In general, we observe that the performance improvement brought by LU is efficient when \#LU is low while the improvement saturates and becomes incremental when \#LU becomes greater. Therefore, the efficient use of computational resources can be realized by limiting the number of LUs to be small to meet the specific requirement at hand.

\subsection{Visual interpretation of LU by curvature analysis}
\label{sec: visual_interpretation}
\begin{figure}[t]
    \centering
    \includegraphics[width=0.95\linewidth]{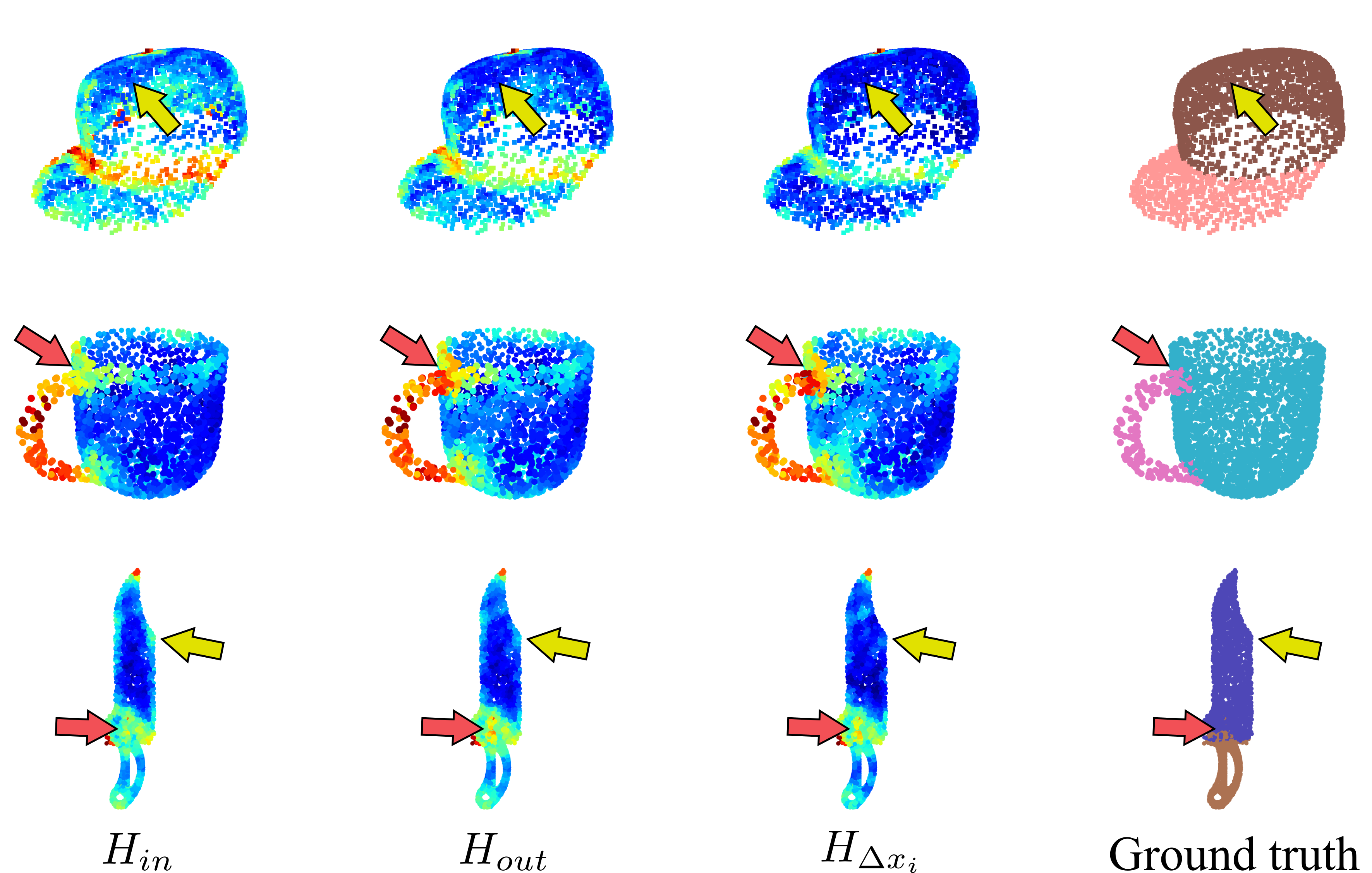}
    \caption{Visualization of curvatures using the ShapeNet Part dataset. \textcolor{yellow}{Yellow} arrows indicate the region smoothed by LU whereas \textcolor{red}{red} arrows point to the sharpened one. }
    \label{fig: curvature_shapenet}
\end{figure}

\begin{figure}[t]
    \centering
    \includegraphics[width=0.98\linewidth]{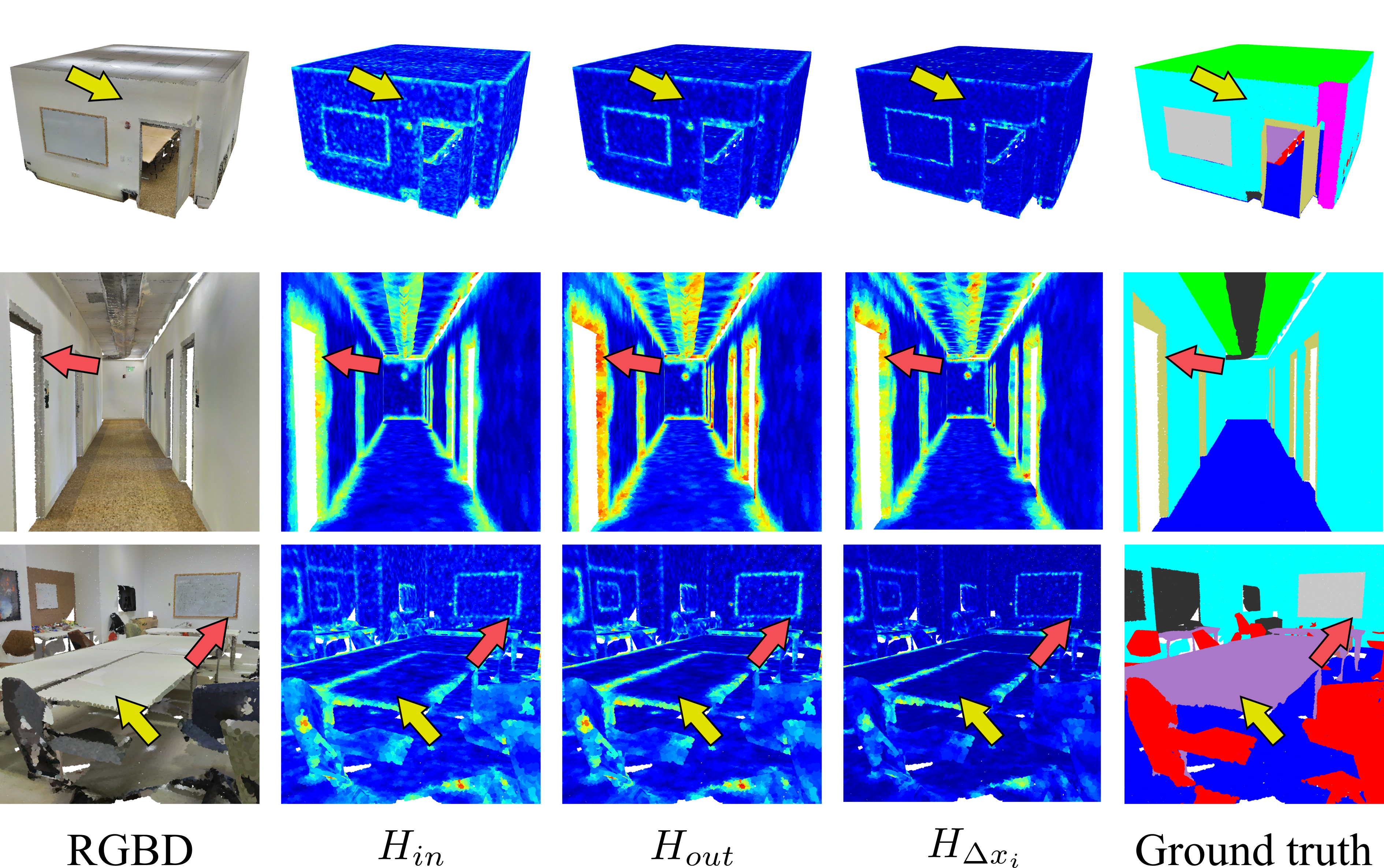}
    \caption{Visualization of curvatures using the S3DIS dataset. \textcolor{yellow}{Yellow} arrows indicate the region smoothed by LU whereas \textcolor{red}{red} arrows point to the enhanced one.}
    \label{fig: curvature_s3dis}
\end{figure}
In this section, we provide qualitative analysis concerning the underlying mechanism of LU and explain how LU improves performance. 
As we describe in Sec.~\ref{sec: interpretation}, the behavior of LU can be investigated by analyzing the change of curvatures. Specifically, we measure the impact of LU by inspecting the change of curvatures qualitatively before and after applying LU. 
Three quantities are used for analysis: the input curvature $H_{in} = ||\sum_{j \in \mathcal{N}(i)} x_j - x_i||$, the output curvature $H_{out} = ||\sum_{j \in \mathcal{N}(i)} x_j'- x_i'||$, and $H_{\Delta x_i}=||\Delta x_i||$. Comparing $H_{in}$ with $H_{out}$ reveals the effect of LU on the smoothness of local surfaces. Furthermore, contrasting $H_{in}$ and $H_{\Delta x_i}$ shows more directly how LU transforms the local features.  
We use the ShapeNet (part segmentation) and S3DIS (scene segmentation) for the analysis. 
The results are shown in Fig.~\ref{fig: curvature_shapenet} and Fig.~\ref{fig: curvature_s3dis}, respectively.  

As indicated by the yellow arrows in Fig.~\ref{fig: curvature_shapenet} and \ref{fig: curvature_s3dis}, LU performs smoothing in some cases just as indicated by its connection to mean curvature flow. As can be seen from the first row of Fig.~\ref{fig: curvature_s3dis}, for instance, small variations on the ceiling and the wall of the room are smoothed. Notice that LU does not blindly smooth out everything; in fact, we observe that the sharp edges of the objects remain salient while within-boundary regions are smoothed. Therefore, LU is able to perform smoothing selectively.

On the other hand, as can be seen from the second rows of Fig.~\ref{fig: curvature_shapenet} and Fig.~\ref{fig: curvature_s3dis}, LU sharpens features by increasing their curvatures in some situations, which is in stark contrast to the mean curvature flow that only performs smoothing. More importantly, LU performs selective sharpening by increasing curvatures for some object edges whereas non-edge regions remain smooth. Furthermore, the fact that most sharpened edges correspond well to the ground truth object boundaries reveals that the effect of LU is task-dependent, further demonstrating its adaptability.

Apart from the cases where LU dominantly performs smoothing or sharpening, we observe that LU can simultaneously perform smoothing and sharpening to different parts of a single point cloud. Such situations are described by the last rows of Fig.~\ref{fig: curvature_shapenet} and Fig.~\ref{fig: curvature_s3dis}. Scrutinizing $H_{in}$ and $H_{out}$ reveals that LU manages to remove small variations for intraregion points while enhancing points near object boundaries. Such an effect is more frequently observed in deeper layers where features are highly semantic.

\section{Discussions}
In designing LU, we especially put emphasis on its lightweightness and optimization friendliness; hence, the design choice adopted in this study is fairly simple and computationally efficient. However, it is highly likely that more sophisticated designs of mappings $\mathcal{M}$ and $\mathcal{T}$ may provide performance improvements at the cost of reduced efficiency. We thus believe that there exists much space to extend its design so that the developed variant can be tailored to a specific kind of point clouds or task. One interesting direction would be the combination of LU and attention mechanism~\citep{vaswani2017attention} in which neighborhood relationship is modeled adaptively. We expect that dynamically constructing the neighborhood graph would result in enhanced robustness against common problems like inconsistent neighborhood or measurement noise.  

Although LU has limited impact on tasks like classification in which globally aggregated information matters, LU provides a significant improvement on more challenging segmentation tasks. Therefore, LU is expected to be rather helpful for tasks involving per-point labelings such as object detection and instance segmentation, which is the direction that we will explore in the future. Furthermore, we believe that LU is especially more influential in applications that require more fine-grained modeling of the local surface geometry. An interesting example of such applications is building damage classification~\citep{xiu2020collapsed} where damage manifested itself as locally deformed surfaces.          

\section{Conclusion}
The fact that a point cloud lacks connectivity information makes it challenging to analyze the geometry of the underlying surface.   
To tackle this issue, this study proposes a simple yet effective architectural unit called Laplacian Unit (LU) that facilitates the learning of local surface geometry.
Observing that the convolution equation consists of coupled modeling of local and global components, LU explicitly decouples their shared optimization and enhances the local feature learning by applying independent transformations to local components. Further, the networks equipped with LUs, namely LU-CNNs, are constructed to tackle point cloud classification, part segmentation, and scene segmentation. Extensive experiments have verified that LU-CNNs achieve competitive or superior performance on several challenging benchmarks. 
In addition, the resulting form of LU enables us to establish straightforward connections between LU and mean curvature flow, an algorithm that smooths the surface using curvature information. We take advantage of such connections and visually interpret the behavior of LU by examining curvatures. As a result, we show by analysis that LU in effect performs adaptive smoothing and sharpening of local surfaces, which leads to improved performance. We believe LU, which explicitly decouples the convolution and enhances the learning of local geometry in an efficient and learning-friendly manner, can be a useful architectural unit for 3D point cloud understanding.

\section*{Acknowledgments}
This work was partially supported by a project commissioned by the New Energy and Industrial Technology Development Organization (JPNP18010), JSPS Grant-in-Aid for Scientific Research (21K12042) and Fundamental Research Funds for the Central Universities under Grant DUT21RC(3)028.

\bibliographystyle{model2-names}
\bibliography{refs}

\end{document}